\documentclass[twoside,11pt]{article}

\usepackage{jmlr2e}
\usepackage{algorithmic}
\usepackage{algorithm}
\usepackage[cmex10]{amsmath}
\usepackage{amssymb}
\usepackage{ifpdf}
\usepackage{rotating}

\usepackage{url}
\usepackage[usenames,dvipsnames]{color}

\newcommand{\eat}[1]{}
\newcommand{\oldversion}[1]{}

\ShortHeadings{The Libra Toolkit for Probabilistic Models}{Lowd and Rooshenas}
\firstpageno{1}

\begin{document}

\title{The Libra Toolkit for Probabilistic Models}

\author{\name Daniel Lowd \email lowd@cs.uoregon.edu \\
       \addr Department of Computer and Information Science\\
       University of Oregon\\
       Eugene, OR 97403, USA
       \AND
       \name Amirmohammad Rooshenas \email pedram@cs.uoregon.edu \\
       \addr Department of Computer and Information Science\\
       University of Oregon\\
       Eugene, OR 97403, USA}

\editor{??}

\maketitle

\begin{abstract}%   <- trailing '%' for backward compatibility of .sty file
The Libra Toolkit is a collection of algorithms for
learning and inference with discrete probabilistic models, including
Bayesian networks, Markov networks, dependency networks, and
sum-product networks.  Compared to other toolkits, Libra places a
greater emphasis on learning the structure of tractable models in
which exact inference is efficient.  It also includes a variety of
algorithms for learning graphical models in which inference is 
potentially intractable, and for performing exact and approximate 
inference.  Libra is released under a 2-clause BSD license
to encourage broad use in academia and industry.
\end{abstract}


\section{Introduction}

The Libra Toolkit is a collection of algorithms for learning and
inference with probabilistic models in discrete domains.  What
distinguishes Libra from other toolkits is the types of methods and
models it supports.  Libra includes a number of algorithms for {\em
structure learning for tractable probabilistic models} in which exact
inference can be done efficiently.  Such models include sum-product
networks (SPN), mixtures of trees (MT), and Bayesian and Markov
networks with compact arithmetic circuits (AC).  These learning
algorithms are not available in any other open-source toolkit.  Libra
also supports {\em structure learning for graphical models}, such as
Bayesian networks (BN), Markov networks (MN), and dependency networks
(DN), in which inference is not necessarily tractable.  Some of these
methods are unique to Libra as well, such as using dependency networks
to learn Markov networks.  Libra provides a variety of exact and
approximate inference algorithms for answering probabilistic queries
in learned or manually specified models.  Many of these are designed
to exploit local structure, such as conjunctive feature functions or
tree-structured conditional probability distributions.

The overall goal of Libra is to make these methods available to
researchers, practitioners, and students for use in experiments,
applications, and education.  Each algorithm in Libra is implemented
in a command-line program suitable for interactive use or scripting,
with consistent options and file formats throughout the toolkit.
Libra also supports the development of new algorithms through modular
code organization, including shared libraries for different
representations and file formats.  

Libra is available under a modified (2-clause) BSD license, which
allows modification and reuse in both academia and industry.

\section{Functionality}

Libra includes a variety of learning and inference algorithms, many of
which are not available in any other open-source toolkit.
See Table~\ref{tab:functionality} for a brief overview.

\newcommand{\pcite}[1]{\protect{\citep{#1}}}

\begin{table}
\begin{center}
\begin{scriptsize}
\begin{tabular}{|lll|}
\hline
\multicolumn{3}{|c|}{Learning General Models} \\
\hline
$\circ$   & BN structure with tree CPDs & \pcite{chickering&al97} \\
          & DN structure with tree/boosted tree/LR CPDs & \pcite{heckerman&al00} \\
$\bullet$ & MN structure from DNs & \pcite{lowd12} \\
          & MN parameters (pseudo-likelihood) & \\
\hline
\multicolumn{3}{|c|}{Learning Tractable Models} \\
\hline
$\bullet$ & Tractable BN/AC structure & \pcite{lowd&domingos08} \\
$\bullet$ & Tractable MN/AC structure & \pcite{lowd&rooshenas13} \\
$\bullet$ & Mixture of trees (MT) & \pcite{meila&jordan00} \\
$\bullet$ & SPN structure (ID-SPN algorithm) & \pcite{rooshenas&lowd14} \\
          %& SPN structure (LearnSPN algorithm) & \pcite{gens&domingos13} \\
          & Chow-Liu algorithm & \pcite{chow&liu68} \\
$\bullet$ & AC parameters (maximum likelihood) & \\
\hline
\multicolumn{3}{|c|}{Approximate Inference} \\
\hline
          & Gibbs sampling (BN,MN,$\bullet$DN) & \pcite{heckerman&al00} (DN) \\
          & Mean field (BN,MN,$\bullet$DN) & \pcite{lowd&shamaei11} (DN) \\
          & Loopy belief propagation (BN,MN) & \\
          & Max-product (BN,MN) & \\
          & Iterated conditional modes (BN,MN,$\bullet$DN) & \\
$\bullet$ & Variational optimization of ACs & \pcite{lowd&domingos10} \\
\hline
\multicolumn{3}{|c|}{Exact Inference} \\
\hline
$\circ$ & AC variable elimination (BN,MN) & \pcite{chavira&darwiche07} \\
$\circ$ & Marginal and MAP inference (AC,SPN,MT) & \pcite{darwiche03} \\
\hline
\end{tabular}
\end{scriptsize}
\end{center}
\caption{Learning and inference algorithms implemented in Libra.
Filled circles ($\bullet$) indicate algorithms that are unique to
Libra, and hollow circles ($\circ$) indicate algorithms with no other
open-source implementation.}
\label{tab:functionality}
\end{table}

Libra's command-line syntax is designed to be simple. For example, to
learn a tractable BN, run the command: ``{\tt libra acbn -i
train.data -mo model.bn -o model.ac}'' where {\tt train.data} is the
input data, {\tt model.bn} is the filename for saving the learned BN,
and {\tt model.ac} is the filename for the corresponding AC
representation, which allows for efficient, exact inference.  To
compute exact conditional marginals in the learned model: ``{\tt libra
acquery -m model.ac -ev test.ev -marg}''.  To compute approximate
marginals in the BN with loopy belief propagation: ``{\tt libra bp -m
model.bn -ev test.ev}''.  Additional command-line parameters can be
used to specify other options, such as the priors and heuristics used
by {\tt acbn} or the maximum number of iterations for {\tt bp}.  These
are just three of more than twenty commands included in Libra.

Libra supports a variety of file formats.  For data instances, Libra
uses comma separated values, where each value is a zero-based index
indicating the discrete value of the corresponding variable.  For
evidence and query files, unknown or missing values are represented
with the special value ``\verb|*|''.  For model files, Libra supports
the XMOD representation from the WinMine Toolkit, the Bayesian
interchange format (BIF), and the simple representation from the UAI
inference competition.  Libra converts among these different formats
using the provided {\tt mconvert} utility, as well as to its own
internal formats for BNs, MNs, and DNs ({\tt .bn}, {\tt .mn}, {\tt
.dn}).  Libra has additional representations for ACs and SPNs ({\tt
.ac}, {\tt .spn}).  These formats are designed to be easy for
humans to read and programs to parse.

Libra is implemented in OCaml.  OCaml is a statically typed language
that supports functional and imperative programming styles, compiles
to native machine code on multiple platforms, and uses type inference
and garbage collection to reduce programmer errors and effort.  OCaml
has a good foreign function interface, which Libra uses for linking to
C libraries and implementing some memory-intensive subroutines more
efficiently in C.  The code to Libra includes nine support libraries,
which provide modules for input, output, and representation of
different types of models, as well as commonly used algorithms and
utility methods.  

\section{Comparison to Other Toolkits}

In Table~\ref{tab:comparison}, we compare Libra to other toolkits in
terms of representation, learning, and inference.

\begin{table}
\begin{center}
\begin{scriptsize}
\begin{tabular}{|l|ll|ll|ll|}
\hline
& \multicolumn{2}{c}{Representation} & \multicolumn{2}{c}{Inference} & \multicolumn{2}{c|}{Learning} \\
Toolkit & Model Types       & Factors          & Exact & Approx. & Param.               & Structure  \\
\hline
Libra                & BN,MN,DN,SPN,AC & Tree,Feature & ACVE & G,BP,MF & ML,PL & BN,\dots,AC \\
\hline              
FastInf              & BN,MN & Table       & JT      & Many & ML,EM & - \\
libDAI               & BN,MN & Table       & JT,E    & Many & ML,EM & - \\
OpenGM2              & BN,MN & Sparse      & -       & Many & -     & - \\
\hline               
Banjo                & BN,DBN    & Table       & -       & -              & -     & BN \\
BNT                  & BN,DBN,ID & LR,OR,NN    & JT,VE,E & G,LW,BP        & ML,EM & BN \\
Deal                 & BN        & Table       & -       & -              & -     & BN \\
OpenMarkov           & BN,MN,ID  & Tree,ADD,OR & JT      & LW             & ML    & BN,MN \\
SMILE                & BN,DBN,ID & Table       & JT      & Sampling       & ML,EM & BN \\ 
UnBBayes             & BN,ID     & Table       & JT      & G,LW           & -     & BN \\
\hline
\end{tabular}
\end{scriptsize}
\end{center}
\caption{Comparison of Libra to several other probabilistic inference and learning toolkits.}
\label{tab:comparison}
\end{table}

In terms of representation, Libra is the only open-source software
package that supports ACs and one of a very small number that support
DNs or SPNs.  Libra does not currently support dynamic Bayesian
networks (DBN) or influence diagrams (ID).  For factors, Libra
supports tables, trees, and arbitrary conjunctive feature functions.
BNT~\citep{murphy01} and OpenMarkov~(CISIAD, 2013)
\nocite{openmarkov13} also support additional types of CPDs, such as
logistic regression, noisy-OR, neural networks, and algebraic decision
diagrams, but they only support tabular CPDs for structure learning.
OpenGM2~\citep{opengm2} supports sparse factors, but iterates through
all factor states during inference.  Libra is unique in its ability to
learn models with local structure and exploit that structure in
inference.

For exact inference, the most common algorithms are junction tree
(JT), enumeration (E), and variable elimination (VE).  Libra provides
ACVE~\citep{chavira&darwiche07}, which is similar to building a
junction tree, but it can exploit structured factors to run inference
in many high-treewidth models.  For approximate inference, Libra
provides Gibbs sampling (G), loopy belief propagation (BP), and mean
field (MF), all of which are optimized for structured factors.  A few
learning toolkits offer likelihood weighting (LW) or additional
sampling algorithms for BNs.  FastInf~\citep{jaimovich&al10},
libDAI~\citep{mooij10}, and OpenGM2 offer the most algorithms but only
support tables.

For learning, Libra supports maximum likelihood (ML) parameter
learning for BNs, ACs, and SPNs, and pseudo-likelihood (PL)
optimization for MNs and DNs.  Libra does not yet support expectation
maximization (EM) for learning with missing values.  Structure
learning is one of Libra's greatest strengths.  Most toolkits only
provide algorithms for learning BNs with tabular CPDs or MNs using the
PC algorithm~\citep{spirtes&al93}.  Libra includes methods for learning
BNs, MNs, DNs, SPNs, and ACs, and all of its algorithms support
learning with local structure.

\begin{figure}
\begin{center}
\includegraphics[scale=0.5]{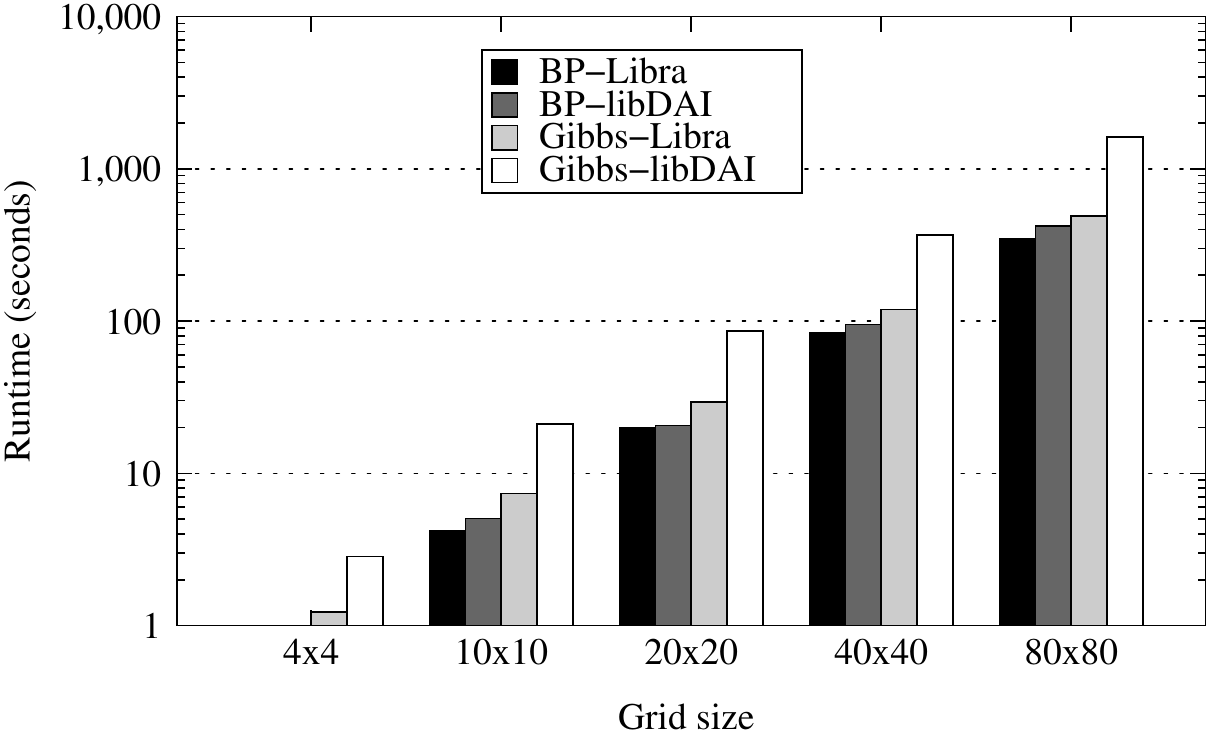}
\vspace{-0.3in}
\end{center}
\caption{Running time of belief propagation and Gibbs sampling in
Libra and libDAI, evaluated on grid-structured MNs of various sizes.}
\label{fig:libdai}
\end{figure}

In experiments on grid-structured MNs, Libra's implementations of BP
and Gibbs sampling were at least as fast as libDAI, a popular C++
implementation of many inference algorithms.  The accuracy of both
toolkits was equivalent.  Parameter settings, such as the number of
iterations, were identical.  See Figure~\ref{fig:libdai} for more
details.

\section{Conclusion}

The Libra Toolkit provides algorithms for learning and inference in a
variety of probabilistic models, including BNs, MNs, DNs, SPNs, and
ACs.  Many of these algorithms are not available in any other
open-source software.  Libra's greatest strength is its support for
tractable probabilistic models, for which very little other software
exists.  Libra makes it easy to use these state-of-the-art methods in
experiments and applications, which we hope will accelerate the
development and deployment of probabilistic methods.

\acks{The development of Libra was partially supported by ARO grant
W911NF-08-1-0242, NSF grant IIS-1118050, NIH grant R01GM103309, 
and a Google Faculty Research Award.  The views and conclusions 
contained in this document are those
of the authors and should not be interpreted as necessarily
representing the official policies, either expressed or implied, of
ARO, NIH, or the United States Government.}

\newcommand{\jmlr}{JMLR}

%\bibliography{all}

\end{document}